\documentclass[runningheads]{llncs}
\usepackage{amsmath}
\usepackage{booktabs}
\usepackage{svg} 
\usepackage{capt-of}   
\usepackage{enumitem}
\usepackage[T1]{fontenc}
\usepackage{graphicx}
\begin{document}
\title{Ev-YOLO: Uncertainty-Aware Object Detection via a Unified Evidential Formulation}
\titlerunning{Ev-YOLO}
%
\author{Simon Barbarit{-}-Gaboriau\inst{1}\orcidID{0009-0008-2977-9552} \and
Hind Laghmara\inst{1}\orcidID{0000-0002-5085-1159} \and
Rémi Boutteau\inst{1}\orcidID{0000-0003-1078-5043} \and
Samia Ainouz\inst{1}\orcidID{0000-0002-2699-4002}}
\authorrunning{S. Barbarit{-}-Gaboriau et al.}
%
\institute{ INSA Rouen Normandie, Univ Rouen Normandie, Univ Le Havre Normandie, \\
Normandie Univ, LITIS UR 4108, F-76000 Rouen, France.
\email{[simon.barbarit-{-}gaboriau@insa-, hind.laghmara@insa-,
remi.boutteau@univ-, samia.ainouz@insa-]rouen.fr}
}
\maketitle              
\begin{abstract}
Reliable uncertainty estimation is essential for deploying object detectors in autonomous systems operating in uncertain environments. Evidential Deep Learning (EDL) provides a principled framework for uncertainty-aware classification by representing network outputs as evidence and interpreting predictions through subjective logic. However, existing evidential object detectors typically combine evidential classification with regression uncertainty models that do not share the same theoretical foundation. In this work, we propose an evidential version of YOLOv8 in which both classification and bounding-box regression are formulated within a common evidential framework. Our approach exploits YOLOv8's distribution-based bounding-box representation, allowing the evidential formulation to be applied not only to classification but also to localisation. As a result, both tasks produce belief, uncertainty, and probability estimates that can be interpreted within the Dempster--Shafer framework. Experiments on KITTI, MUSES, and nuScenes show that the resulting detector remains broadly competitive with standard YOLOv8 in terms of detection accuracy while providing a localisation uncertainty that effectively discriminates between correct and erroneous detections. Moreover, this uncertainty becomes increasingly discriminative under domain shift.


\end{abstract}
\section{Introduction}
Modern object detectors such as YOLOv8 achieve remarkable accuracy and real-time performance, making them attractive for robotics and autonomous driving \cite{Jocher_Ultralytics_YOLO26_Unified_2026}. However, their predictions are often overconfident, especially when operating in conditions that differ from the training data. In safety-critical applications, it is important not only to detect objects accurately but also to quantify the uncertainty associated with each prediction.

Evidential Deep Learning (EDL) has recently emerged as an efficient, single-pass approach for uncertainty estimation. Rather than directly predicting probabilities, EDL predicts evidence and derives belief and uncertainty measures through subjective logic \cite{sensoy2018evidential,josang2016subjective}. While several works have extended EDL to object detection, most focus on classification uncertainty \cite{park2023active}, and those that address localisation typically rely on Deep Evidential Regression (DER) proposed by Amini \emph{et al.}\cite{amini2020deep}. Although both approaches are often referred to as evidential, they rely on fundamentally different mathematical formulations. Sensoy's method \cite{sensoy2018evidential} produces Dirichlet opinions that can be directly interpreted through subjective logic and Dempster--Shafer theory, whereas DER models regression targets using Normal-Inverse-Gamma distributions that do not define belief masses. Classification and localisation uncertainties therefore live in different frameworks, preventing a unified interpretation of detector outputs.

In this paper, we propose Ev-YOLO, a unified evidential formulation for object detectors based on Distribution Focal Loss (DFL), and demonstrate its implementation on YOLOv8. Our key observation is that DFL represents bounding-box coordinates as discrete distributions over bins rather than as direct regression targets \cite{li2020generalized}. This representation naturally matches the Dirichlet-based formulation introduced by Sensoy \emph{et al.} \cite{sensoy2018evidential}, enabling evidential learning to be applied directly to localisation. Consequently, any object detector relying on DFL can benefit from the proposed formulation. We reformulate both regression and classification within the same subjective logic framework, yielding belief, uncertainty, and probability estimates for localisation and classification. For classification, we preserve the multi-label design employed by YOLOv8 and many modern detectors by modeling each class independently using evidential binary classifiers.

Experiments on KITTI \cite{Geiger2012CVPR}, MUSES \cite{brodermann2024muses}, and nuScenes \cite{nuscenes2019} show that the proposed detector remains competitive in detection accuracy while being better calibrated and providing a localisation uncertainty that reliably identifies erroneous detections and becomes more discriminative under domain shift.

The main contributions of this work are:

\begin{itemize}[topsep=0pt]
\item A unified evidential formulation of YOLOv8 for both classification and bounding box regression.
\item An evidential reinterpretation of Distribution Focal Loss enabling belief and uncertainty estimation for localisation.
\item A localisation uncertainty that reliably discriminates correct from erroneous detections and becomes more discriminative under domain shift, while detection accuracy remains competitive.
\end{itemize}

\section{Evidential Object Detection using Dempster--Shafer Theory and Subjective Logic}

To explicitly model predictive uncertainty in object detection, we replace the probabilistic outputs of the standard YOLOv8 detection head by evidential representations derived from the Dempster--Shafer Theory (DST) of evidence. In particular, evidential reasoning is applied both to bounding box localisation and object classification.

The original YOLOv8 detector is optimized using three loss components: a localisation loss based on Complete IoU (CIoU), a DFL for bounding box regression, and a Binary Cross-Entropy (BCE) loss for classification. In our evidential formulation, the localisation loss remains unchanged, while the DFL and classification losses are replaced by evidential objectives derived from the framework of Sensoy et al. The overall training objective is

\begin{equation}
\mathcal{L}
= \lambda_{\mathrm{box}} \mathcal{L}_{\mathrm{CIoU}} +\lambda_{\mathrm{dfl}} \mathcal{L}_{\mathrm{E-DFL}}+\lambda_{\mathrm{cls}} \mathcal{L}_{\mathrm{E-CLS}},
\end{equation}

where $\mathcal{L}_{\mathrm{CIoU}}$ denotes the standard YOLOv8 bounding box loss, $\mathcal{L}_{\mathrm{E-DFL}}$ the evidential localisation loss, and $\mathcal{L}_{\mathrm{E-CLS}}$ the evidential classification loss.

\subsection{Evidential Bounding Box Regression}

In the original YOLOv8 architecture, each distance from the anchor to the edge of a bounding box is represented by a discrete probability distribution over $K$ bins (typically $K=16$). Let $\mathbf{z}_i = [z_1,\ldots,z_K]$ denote the raw logits associated with the distance $i \in \{t, b, l, r\}$ for the top, bottom, left and right edges of a bounding box. The standard formulation converts these logits into probabilities using a softmax function. Instead, we adopt the evidential framework first proposed by Sensoy et al. \cite{sensoy2018evidential}, based on subjective logic, and interpret the network outputs as evidence supporting each bin. For each distance $i \in \{t, b, l, r\}$, the logits are first transformed into non-negative evidence values using the Softplus activation function: $e_{i, k} = \mathrm{Softplus}(z_{i, k}).$ The evidence vector $\mathbf{e}$ parameterizes a Dirichlet distribution through

\begin{equation}
\alpha_{i,k} = e_{i,k} + 1,
\qquad
S_i = \sum_{k=1}^{K} \alpha_{i,k}.
\end{equation}

Following the subjective logic interpretation of the Dirichlet distribution:

\begin{equation}
b_{i,k} = \frac{e_{i,k}}{S_i},
\qquad
u_i = \frac{K}{S_i},
\qquad
p_{i,k} 
= \frac{\alpha_{i,k}}{S_i},
\qquad
\hat{x}_i = \sum_{k=0}^{K-1} kp_{i,k}.
\end{equation}

With $b_{i,k}$ the belief mass assigned to bin $k$, $u_i$ the uncertainty mass, $p_{i,k}$ the projected probability associated with each bin and $\hat{x}_i$ the expected coordinate value computed identically to DFL. The uncertainty $u_i$ is a measure correlated to epistemic uncertainty: large values indicate insufficient evidence to select a bin, while small values correspond to confident predictions.

Let $y_i$ be the target coordinate value and let $l_i=\lfloor y_i \rfloor$ and $r_i=l_i+1$ denote the neighboring integer bins. The original DFL objective supervises both bins using a weighted cross-entropy loss,

\begin{equation}
\mathcal{L}_{\mathrm{DFL}}^{(i)}
=
(y_i-l_i)\mathrm{CE}(z_i, r_i)
+
(r_i-y_i)\mathrm{CE}(z_i, l_i).
\end{equation}

With this loss being applied to each distance $i \in \{t, b, l, r\}$ from the anchor to the box edges. To obtain an evidential formulation, we replace the general cross-entropy term by the evidential classification loss.


\begin{equation}
\mathcal{L}_{\mathrm{E-CE}}= \sum_{k=1}^{K} y_k(\psi(S) - \psi(\alpha_k)),
\end{equation}

where $\psi(\cdot)$ is the digamma function. To discourage unwarranted evidence, a Kullback--Leibler regularization term is added and the evidence supporting incorrect classes is removed:

\begin{equation}
\begin{gathered}
\tilde{\alpha}_k=y_k+(1-y_k)\alpha_k, \qquad \mathcal{L}_{\mathrm{KL}}=\mathrm{KL}\Big(D(p, \tilde{\alpha})||D(p, 1)\Big) \\
\mathcal{L}_{\mathrm{KL}}= \log\Big(\frac{\Gamma(\sum_{k=1}^{K}\tilde{\boldsymbol{\alpha}}_k)}{\Gamma(K)\Pi_{k=1}^{K}\Gamma(\tilde{\boldsymbol{\alpha}}_k)}\Big)+\sum_{k=1}^{K}(\tilde{\boldsymbol{\alpha}}_k-1)[\psi(\tilde{\boldsymbol{\alpha}}_k)-\psi(\sum_{k=1}^{K}\tilde{\boldsymbol{\alpha}}_k)]
\end{gathered}
\end{equation}

where $D(\cdot)$ denotes a Dirichlet distribution and $D(p, 1)$ is the uniform Dirichlet prior. The evidential loss associated with a target bin becomes

\begin{equation}
\mathcal{L}_{\mathrm{Evi}}=
\mathcal{L}_{\mathrm{E-CE}}
+
\lambda_t
\mathcal{L}_{\mathrm{KL}},
\qquad
\lambda_t=\min\!\left(0.1,\,(\mathrm{epoch}/10)\cdot 0.1\right)
\end{equation}

Unlike the classification head, the localisation KL weight is capped at $0.1$ rather than annealed to $1$. At full weight the regulariser flattens the coordinate distribution faster than the evidence loss can sharpen it, causing training to collapse. The final evidential DFL objective is obtained by replacing each cross-entropy term of the original DFL with the corresponding evidential loss,

\begin{equation}
\mathcal{L}_{\mathrm{E-DFL}}^{(i)}
= (y_i-l_i)
\mathcal{L}_{\mathrm{Evi}}^{(i)}(\alpha_i, r_i)
+
(r_i-y_i)
\mathcal{L}_{\mathrm{Evi}}^{(i)}(\alpha_i, l_i).
\end{equation}

This loss preserves the interpolation mechanism of DFL while enabling uncertainty-aware learning through the Dirichlet evidence representation.

\subsection{Evidential Classification}

For multi-label object classification, YOLOv8 employs independent sigmoid activations for each class. Let $z_c$ denote the logit associated with class $c$. The standard confidence score is $p_c = \sigma(z_c).$

To obtain an evidential representation, each class prediction is modeled as a binary hypothesis corresponding to the presence or absence of the class, similarly to \cite{yuan2020evidential}, with the model producing two logits per class. Following the evidential binary classification framework, the evidence for class $c$ is obtained as

\begin{equation}
e_c = \mathrm{Softplus}(z_c),
\qquad
\hat{e}_c = \mathrm{Softplus}(\hat{z}_c),
\qquad
\alpha_c = e_c + 1,
\qquad
\beta_c = \hat{e}_c + 1.
\end{equation}

With $e_c$ and $\hat{e}_c$ the positive evidence and negative evidence that define the parameters $\alpha_c$ and $\beta_c$ of a Beta distribution. Using the subjective logic formulation, we get:

\begin{equation}
b_c = \frac{e_c}{\alpha_c + \beta_c},
\qquad
d_c = \frac{\hat{e}_c}{\alpha_c + \beta_c},
\qquad
u_c = \frac{2}{\alpha_c + \beta_c},
\qquad
p_c = \frac{\alpha_c}{\alpha_c+\beta_c}.
\end{equation}

With $b_c$ and $d_c$ the belief and disbelief the model has in class $c$ being the answer and $u_c$ and $p_c$ the uncertainty and predictive probability.

As in the regression case, uncertainty is inversely proportional to the accumulated evidence. Predictions supported by little evidence produce large uncertainty values.

For classification, YOLOv8 employs an independent Binary Cross-Entropy loss for each class. We replace this objective with an evidential binary classification loss based on Beta distributions \cite{ashfaq2023deed}. Given a binary target $y_c \in \{0,1\}$, the evidential classification loss is

\begin{equation}
\mathcal{L}_{\mathrm{E-BCE}}^{(c)}
=
y_c(\psi(\alpha_c + \beta_c) - \psi(\alpha_c)) + (1-y_c)(\psi(\alpha_c + \beta_c) - \psi(\beta_c)).
\end{equation}

As in the regression case, a KL regularization term is introduced to suppress unsupported evidence, with $\tilde{\alpha}_c=y_c+(1-y_c)\alpha_c$ and $\tilde{\beta}_c=(1-y_c)+y_c\beta_c$, we have:

\begin{equation}
\mathcal{L}_{\mathrm{KL}}^{(c)}=
\mathrm{KL}
\Big(
\mathrm{Beta}
(\tilde{\alpha}_c,\tilde{\beta}_c)
||
\mathrm{Beta}(1,1)
\Big),
\end{equation}
\begin{equation}
\mathcal{L}_{\mathrm{KL}}^{(c)}=
\log \Big(\frac{\Gamma(\tilde{\alpha}_c + \tilde{\beta}_c)}{\Gamma(\tilde{\alpha}_c)\Gamma(\tilde{\beta}_c)}\Big)
+ (\tilde{\alpha}_c-1)\psi(\tilde{\alpha}_c) + (\tilde{\beta}_c - 1)\psi(\tilde{\beta}_c)-(\tilde{\alpha}_c + \tilde{\beta}_c -2)\psi(\tilde{\alpha}_c + \tilde{\beta}_c).
\end{equation}

The evidential loss for class $c$ is therefore

\begin{equation}
\mathcal{L}_{\mathrm{Evi}}^{(c)}
=
\mathcal{L}_{\mathrm{E-BCE}}^{(c)}
+
\lambda_t
\mathcal{L}_{\mathrm{KL}}^{(c)},
\qquad
\lambda_t=\min\left(1,\mathrm{epoch}/10\right)
\end{equation}

The final classification loss is obtained by averaging over all classes.

Compared with the standard BCE objective, this formulation encourages the network to distinguish between prediction errors and lack of evidence, producing calibrated class probabilities together with explicit uncertainty estimates.

\section{Results}
The baseline YOLOv8n and our evidential variant are trained under identical conditions on a single NVIDIA H200 GPU. Both models are trained for 200 epochs using AdamW with an initial learning rate of $1.667 \times 10^{-3}$ and $\beta_1 = 0.9$, batch size 16 and input resolution $640\times640$. The evidential regression head uses $K=16$ discretisation bins. All hyperparameters, optimisation settings, and the random seed are shared; models differ only in the detection head and loss.

Both models are evaluated without fine-tuning on the in-domain KITTI evaluation split (1{,}000 images; the remaining 6{,}481 are used for training) and, to probe robustness under distribution shift, on the MUSES and nuScenes datasets, with MUSES being a particularly severe shift due to its adverse meteorological conditions. To avoid label inconsistencies across datasets, all evaluations are restricted to the \emph{car} and \emph{pedestrian} classes.

\begin{table}[h!]
\centering 
  \begin{minipage}[c]{0.44\textwidth}
    \centering
    \caption{Results of an evaluation on KITTI, MUSES and nuScenes for a model trained on KITTI, presented using mAP$_{50}$.}
    \label{tab:map} 
    \setlength{\tabcolsep}{1pt} 
    \begin{tabular}{l|c|c|c}
    \toprule
    \textbf{Model} & \textbf{KITTI} & \textbf{MUSES} & \textbf{nuScenes} \\
    \midrule
    YOLOv8n & \textbf{90.95} & 6.05 & \textbf{17.80} \\
    \midrule
    Ours & 86.30 & \textbf{6.70} & 17.32 \\
    \bottomrule
    \end{tabular}
  \end{minipage}
  \hfill 
  \begin{minipage}[c]{0.52\textwidth}
    \centering
    \caption{Results of an evaluation on KITTI, MUSES and nuScenes for a model trained on KITTI with the AUROC metric on localisation.}
    \label{tab:auroc}
    \setlength{\tabcolsep}{1pt} 
    \begin{tabular}{l|c|c|c}
    \toprule
    \textbf{Model} & \textbf{KITTI} & \textbf{MUSES} & \textbf{nuScenes} \\
    \midrule
    YOLOv8n($H$) & 79.08 & 82.76 & 88.45 \\
    \midrule
    Ours($H$) & 94.43 & 98.17 & 97.93 \\
    \midrule
    Ours($K/S$) & 93.09 & 95.32 & 93.48 \\
    \bottomrule
    \end{tabular}
  \end{minipage}
\end{table}

Table~\ref{tab:map} reports mAP$_{50}$. The proposed model remains competitive with the baseline both in-domain and under shift, despite a moderate reduction in in-domain mAP. As detection accuracy is not the primary objective of the evidential formulation, these results primarily confirm that incorporating localisation uncertainty does not degrade nominal performance.

The main contribution lies in applying evidential learning to Distribution Focal Loss, thereby equipping each predicted box edge with calibrated localisation uncertainty. This is assessed in Table~\ref{tab:auroc} via AUROC between the predicted uncertainty and the correctness label (IoU threshold $0.5$), which quantifies the probability that an incorrect detection receives higher uncertainty than a correct one. For fair comparison, both models are first evaluated using predictive entropy $H$ over the per-coordinate bin distributions. Under this metric, the evidential model consistently achieves superior error discrimination across all datasets, with discrimination improving rather than degrading under distribution shift, consistent with an uncertainty that grows under unfamiliar conditions. Replacing entropy with evidential uncertainty $K/S$ yields comparable results, indicating that the improvement arises from the evidential formulation itself rather than a specific uncertainty estimator.



\begin{figure}[t]
\centering
  \begin{minipage}[c]{0.46\textwidth}
    \centering
    \setlength{\tabcolsep}{2pt}
    \begin{tabular}{l|c|c|c}
    \toprule
    \textbf{Model} & \textbf{KITTI} & \textbf{MUSES} & \textbf{nuScenes} \\
    \midrule
    YOLOv8n & 0.0335 & 0.1539 & 0.1882 \\
    \midrule
    Ours & \textbf{0.0051} & \textbf{0.0012} & \textbf{0.0015} \\
    \bottomrule
    \end{tabular}
    \captionof{table}{Expected Calibration Error (ECE \cite{kuppers2020multivariate}) on KITTI, MUSES, and nuScenes.}
    \label{tab:ece}
  \end{minipage}
  \hfill
  \begin{minipage}[c]{0.48\textwidth}
    \centering
    \includegraphics[width=\textwidth]{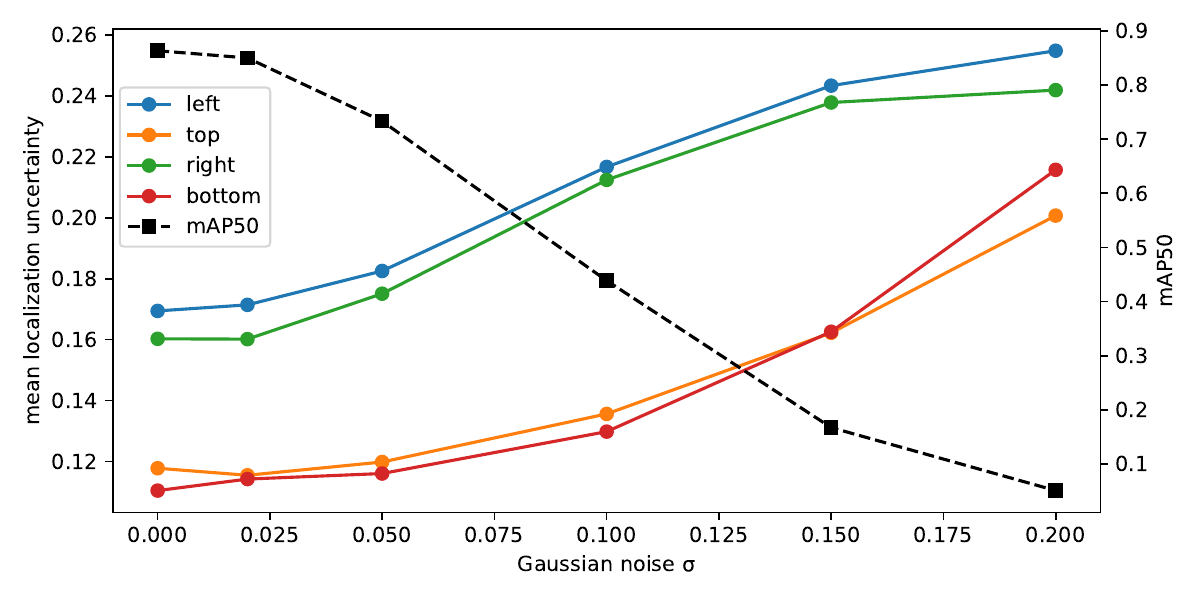}
    \caption{Mean localisation uncertainty and $\mathrm{mAP}_{50}$ against Gaussian
      noise std on KITTI.}
    \label{fig:noise}
  \end{minipage}
\end{figure}

Robustness to input perturbations is further evaluated by corrupting KITTI images with additive Gaussian noise and measuring mean localisation uncertainty at ground-truth object locations. As shown in Figure~\ref{fig:noise}, uncertainty increases monotonically as mAP$_{50}$ decreases, confirming sensitivity to genuine input degradation rather than dataset-specific artefacts.

For completeness we also examine the calibration of the classification confidence (Table~\ref{tab:ece}) through Expected Calibration Error (ECE \cite{kuppers2020multivariate}). The proposed model exhibits a lower ECE in-domain, and the gap widens markedly under shift, where the baseline grows substantially over-confident. This result should be interpreted with caution: the baseline is not recalibrated; however, post-hoc calibration on in-distribution data is known to transfer poorly under shift \cite{ovadia2019can}, so this is unlikely to be a decisive disadvantage. The evidential formulation also induces a mild under-confidence bias that mechanically lowers ECE, so part of the improvement reflects this effect; we therefore present it as supporting rather than primary evidence.

Overall, these results show that an evidential DFL head furnishes a localisation uncertainty that is both a strong error detector and demonstrably responsive to distribution shift and input corruption, while leaving detection accuracy essentially unchanged and improving classification calibration.

\section{Conclusion}
We presented a unified evidential formulation of object detection in which localisation and classification are expressed within the same subjective-logic and Dempster--Shafer framework. The key observation is that Distribution Focal Loss already encodes each box coordinate as a discrete distribution over bins, naturally mapping to a Dirichlet evidence model; this extends evidential treatment from classification to localisation without architectural changes, enabling any DFL-based detector to adopt it. On YOLOv8, the resulting detector maintains competitive accuracy while improving calibration and providing informative localisation uncertainty: it better separates correct from erroneous detections than an entropy baseline, becomes more discriminative under domain shift, and increases monotonically as input quality degrades.

A central contribution is the shared evidential representation for localisation and classification, where both outputs are expressed as belief masses. This opens the door to future work on Dempster--Shafer fusion across time, overlapping detections, or sensing modalities, with explicit uncertainty propagation, as well as potential applications to active learning and open-world detection.

Limitations include preserved rather than improved accuracy under domain shift, where gains may partly stem from evidential loss regularisation rather than the representation itself. Classification uncertainty is also less robust than localisation, degrading under synthetic corruption as expected in single-pass evidential models. Future work should further disentangle aleatoric and epistemic localisation uncertainty and validate the approach on additional DFL-based detectors.

\begin{credits}
\subsubsection{\ackname} The first author is supported by the French National Research Agency (ANR) under grant AdaV. This project was provided with computing resources of CRIANN (Normandy, France).

\subsubsection{\discintname}
The authors have no competing interests to declare that are relevant to the content
of this article.
\end{credits}
%
%
%
\bibliographystyle{splncs04}
\bibliography{biblio}

\begin{thebibliography}{10}
\providecommand{\url}[1]{\texttt{#1}}
\providecommand{\urlprefix}{URL }
\providecommand{\doi}[1]{https://doi.org/#1}

\bibitem{amini2020deep}
Amini, A., Schwarting, W., Soleimany, A., Rus, D.: Deep evidential regression.
  Advances in neural information processing systems  \textbf{33},  14927--14937
  (2020)

\bibitem{ashfaq2023deed}
Ashfaq, A., Lingman, M., Sensoy, M., Nowaczyk, S.: Deed: Deep evidential
  doctor. Artificial Intelligence  \textbf{325},  104019 (2023)

\bibitem{brodermann2024muses}
Br{\"o}dermann, T., Bruggemann, D., Sakaridis, C., Ta, K., Liagouris, O.,
  Corkill, J., Van~Gool, L.: Muses: The multi-sensor semantic perception
  dataset for driving under uncertainty. In: European Conference on Computer
  Vision. pp. 21--38. Springer (2024)

\bibitem{nuscenes2019}
Caesar, H., Bankiti, V., Lang, A.H., Vora, S., Liong, V.E., Xu, Q., Krishnan,
  A., Pan, Y., Baldan, G., Beijbom, O.: nuscenes: A multimodal dataset for
  autonomous driving. In: Proceedings of the IEEE/CVF conference on computer
  vision and pattern recognition. pp. 11621--11631 (2020)

\bibitem{Geiger2012CVPR}
Geiger, A., Lenz, P., Urtasun, R.: Are we ready for autonomous driving? the
  kitti vision benchmark suite. In: Conference on Computer Vision and Pattern
  Recognition (CVPR) (2012)

\bibitem{Jocher_Ultralytics_YOLO26_Unified_2026}
Jocher, G., Qiu, J., Liu, M., Lyu, S., Akyon, F.C., Kalfaoglu, M.E.:
  {Ultralytics YOLO26: Unified Real-Time End-to-End Vision Models}  (2026)

\bibitem{josang2016subjective}
J{\o}sang, A.: Subjective logic, vol.~3. Springer (2016)

\bibitem{kuppers2020multivariate}
Kuppers, F., Kronenberger, J., Shantia, A., Haselhoff, A.: Multivariate
  confidence calibration for object detection. In: Proceedings of the IEEE/CVF
  conference on computer vision and pattern recognition workshops. pp. 326--327
  (2020)

\bibitem{li2020generalized}
Li, X., Wang, W., Wu, L., Chen, S., Hu, X., Li, J., Tang, J., Yang, J.:
  Generalized focal loss: Learning qualified and distributed bounding boxes for
  dense object detection. Advances in neural information processing systems
  \textbf{33},  21002--21012 (2020)

\bibitem{ovadia2019can}
Ovadia, Y., Fertig, E., Ren, J., Nado, Z., Sculley, D., Nowozin, S., Dillon,
  J., Lakshminarayanan, B., Snoek, J.: Can you trust your model's uncertainty?
  evaluating predictive uncertainty under dataset shift. Advances in neural
  information processing systems  \textbf{32} (2019)

\bibitem{park2023active}
Park, Y., Choi, W., Kim, S., Han, D.J., Moon, J.: Active learning for object
  detection with evidential deep learning and hierarchical uncertainty
  aggregation. In: The Eleventh International Conference on Learning
  Representations (2023)

\bibitem{sensoy2018evidential}
Sensoy, M., Kaplan, L., Kandemir, M.: Evidential deep learning to quantify
  classification uncertainty. Advances in neural information processing systems
   \textbf{31} (2018)

\bibitem{yuan2020evidential}
Yuan, B., Yue, X., Lv, Y., Denoeux, T.: Evidential deep neural networks for
  uncertain data classification. In: International conference on knowledge
  science, engineering and management. pp. 427--437. Springer (2020)

\end{thebibliography}




\end{document}